\documentclass[journal,twoside]{IEEEtran}
\usepackage{verbatim}

\usepackage{cite}

\ifCLASSINFOpdf
  \usepackage[pdftex]{graphicx}
\else
\fi
\usepackage[linesnumbered, ruled]{algorithm2e}

\usepackage{array}
\usepackage{multirow}
\begin{document}
%
\title{IR$^2$Net: Information Restriction and Information Recovery for Accurate Binary Neural Networks}
%
%
%

\author{Ping Xue,
        Yang Lu,  
        Jingfei Chang, 
        Xing Wei,
        and Zhen Wei      
\thanks{This work was supported in part by the National Key Research and Development Program under Grant 2018YFC0604404, in part by the National Natural Science Foundation of China under Grant 61806067, in part by the Anhui Provincial Key R\&D Program (202004a05020040), and in part by the Intelligent Network and New Energy Vehicle Special Project of Intelligent Manufacturing Institute of HFUT (IMIWL2019003). (\textit{Corresponding author: Yang Lu}.)}
\thanks{Ping Xue and Jingfei Chang are with the School of Computer Science and Information Engineering, 
	Hefei University of Technology, Hefei 230009, China 
	(e-mail: xueping1001@126.com; cjfhfut@mail.hfut.edu.cn).}
\thanks{Yang Lu and Zhen Wei are with the School of Computer Science and Information Engineering, 
	Hefei University of Technology, Hefei 230009, China, with the Anhui Mine IOT and Security Monitoring Technology Key Laboratory, Hefei 230088, China , and also with the Engineering Research Center of Safety Critical Industrial Measurement and Control Technology, Ministry of Education, Hefei University of Technology, Hefei 230009, China (e-mail: luyang.hf@126.com; weizhen@gocom.cn).}
\thanks{Xing Wei is with the School of Computer Science and Information Engineering, 
	Hefei University of Technology, Hefei 230009, China, and also with the Intelligent Manufacturing Institute of HeFei University of Technology, Hefei 230009, China
	(e-mail: weixing@hfut.edu.cn).}}

%
%

\markboth{Preprint}
{Ping Xue \MakeLowercase{\textit{et al.}}: IR$^2$Net: Information Restriction and Information Recovery for Accurate Binary Neural Networks}
%



\maketitle

\begin{abstract}
Weight and activation binarization can efficiently compress deep neural networks and accelerate model inference, but cause severe accuracy degradation. Existing optimization methods for binary neural networks (BNNs) focus on fitting full-precision networks to reduce quantization errors, and suffer from the trade-off between accuracy and computational complexity. In contrast, considering the limited learning ability and information loss caused by the limited representational capability of BNNs, we propose IR$^2$Net to stimulate the potential of BNNs and improve the network accuracy by restricting the input information and recovering the feature information, including: 1) information restriction: for a BNN, by evaluating the learning ability on the input information, discarding some of the information it cannot focus on, and limiting the amount of input information to match its learning ability; 2) information recovery: due to the information loss in forward propagation, the output feature information of the network is not enough to support accurate classification. By selecting some shallow feature maps with richer information, and fusing them with the final feature maps to recover the feature information. In addition, the computational cost is reduced by streamlining the information recovery method to strike a better trade-off between accuracy and efficiency. Experimental results demonstrate that our approach still achieves comparable accuracy even with $ \sim $10x floating-point operations (FLOPs) reduction for ResNet-18. The models and code are available at https://github.com/pingxue-hfut/IR2Net.
\end{abstract}


%
\IEEEpeerreviewmaketitle

\section{Introduction}
%
%
%
%
\IEEEPARstart{D}{eep} Convolutional Neural Networks (CNNs) have made much progress in a wide variety of computer vision applications \cite{1:nips/KrizhevskySH12,2:iccv/HeGDG17,3:cvpr/LongSD15,49:tip/DingMWXCSWL21}. However, as the research advances, the depth of the networks has expanded from a few layers to hundreds of layers \cite{4:726791,5:iclr/SimonyanZ14a,6:cvpr/SzegedyLJSRAEVR15,7:cvpr/HeZRS16}. The huge number of parameters and the ultra-high computational complexity of CNNs make their deployment very constrained, especially under the conditions of applications with high real-time requirements or limited storage capacity. To solve this problem, various compression techniques for CNNs have emerged. Network pruning \cite{8:iccv/HeZS17,9:icme/GeLZJZ17,10:tip/DingZJZH21} reduces model redundancy by pruning convolutional kernels or channels, efficient architecture design \cite{11:corr/HowardZCKWWAA17,12:cvpr/ZhangZLS18,13:cvpr/HanW0GXX20} replaces conventional convolutional layers with well-designed lightweight modules to speed up network inference, knowledge distillation \cite{14:corr/HintonVD15,15:iccv/HouMLL19} attempts to transfer knowledge from complex networks (teachers) to compact networks (students), quantization \cite{16:icml/GuptaAGN15,17:cvpr/WuLWHC16,18:jmlr/HubaraCSEB17,19:aaai/LengDLZJ18,20:iclr/ZhouYGXC17,21:iccv/GongLJLHLYY19} replaces 32-bit weights and activations with low-bit (e.g., 16-bit) ones to reduce both memory footprint and computational complexity. The extreme of quantization is binarization. Compared with 32-bit floating-point networks, network binarization constrains both the weights and activations to \{-1, +1\}, i.e., the parameters of binary neural networks (BNNs) need only 1-bit representation, which greatly reduces the storage requirement; furthermore, while binarizing the network weights and activations, the computationally intensive matrix multiplication and addition operations in full-precision networks are replaced with low-cost XNOR and bitcount, which greatly reduces the network inference delay. Therefore, benefiting from the high compression ratio, acceleration, and energy-saving, network binarization is considered as one of the most promising techniques for network compression and is the focus of this work.\\
\begin{figure}[!t]
	\centering
	\includegraphics[width=3.5in]{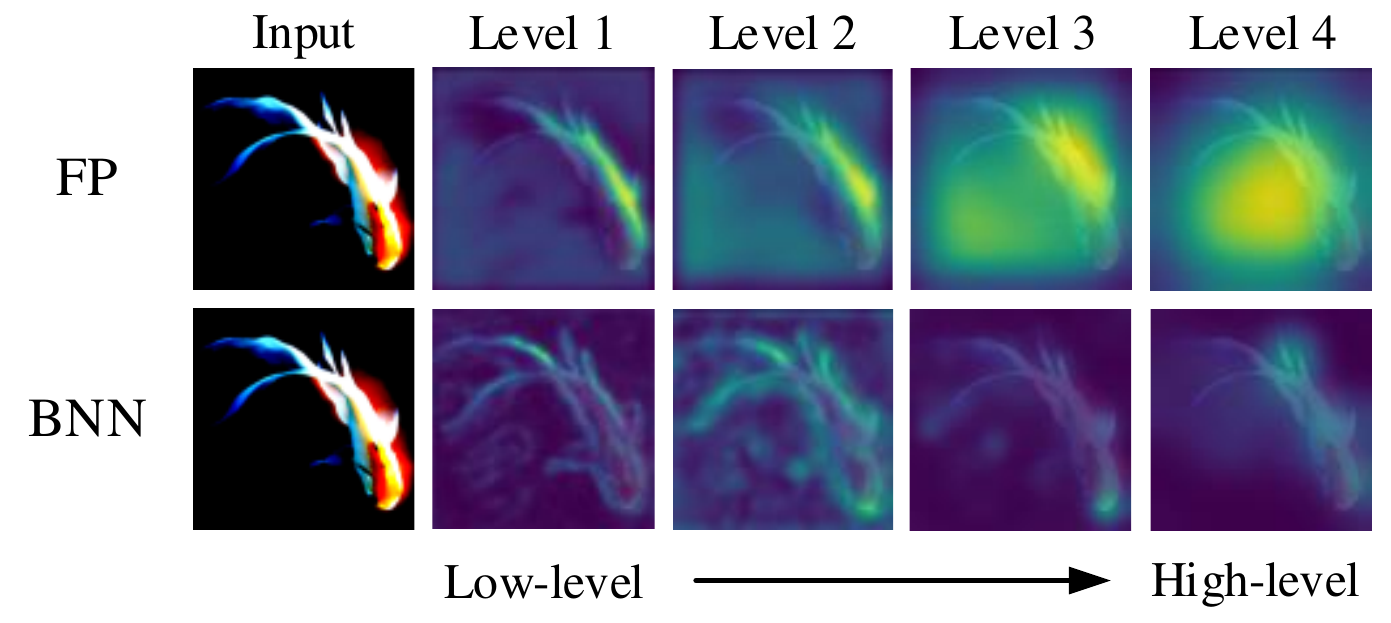}
	\caption{The differences of attention maps for full-precision network (FP) and BNN.}
\end{figure}
\indent
Network binarization has attracted a lot of attention due to its advantages in compression and acceleration. Although much progress has been made, the existing binarization methods still suffer from a trade-off between accuracy and efficiency. For example, XNOR-Net \cite{22:eccv/RastegariORF16} and Bi-Real Net \cite{23:eccv/LiuWLYLC18} have improved the accuracy of BNNs with negligible extra computation, there remains a large accuracy gap between them and the full-precision counterparts; whereas Group-Net \cite{24:cvpr/ZhuangSTL019} and MeliusNet \cite{25:corr/abs-2001-05936} achieve comparable accuracy to that of full-precision networks, but they introduce a noticeable additional computational cost, which significantly offsets the advantages of network binarization. Therefore, one of the motivations for this work is to strike a better trade-off between the accuracy and computational complexity for BNNs.\\
\indent
In addition, the performance degradation of BNNs is mainly caused by their limited representational capability. BNNs represent weights and activations with 1-bit, which means the theoretical representation precision is only 1/2$^{31}$ compared to the full-precision counterparts. The limited representational capability leads to two drawbacks in BNNs: limited data information acceptance (i.e., learning ability) and severe information loss during forward propagation. As shown in Figure 1, at level 4 of the attention maps \cite{26:iclr/ZagoruykoK17}, it can be seen that the full-precision network can focus on much larger information regions of interest (the highlighted regions of the attention maps) than the BNN do, which is only able to accept limited information; besides, the information loss during the forward propagation of the BNN is also evident in the flow of the attention maps from low to high levels. IR-Net \cite{27:cvpr/QinGLSWYS20} and BBG \cite{28:icassp/ShenLGH20} reduce the information loss in forward propagation by balancing and normalizing the weights to achieve maximum information entropy, which improves the network accuracy to some extent. However, these methods do not consider the limited information acceptance of BNNs, while they remain significant accuracy degradation on large-scale datasets (e.g., ImageNet).\\
\indent
To solve the aforementioned problems, from the perspective of the representational capability of BNNs themselves, we propose IR$^2$Net, a binarization approach to enhance BNNs via restricting input information and recovering feature information: 1) intuitively, different students (networks) have different learning abilities, for those with strong learning abilities, more information can be provided for their learning and refining, whereas for those with weak learning abilities, discarding redundant information is needed for better learning. IR$^2$Net introduces the information restriction method to restrict the input information and regularize the networks, thus forces BNNs to focus on the more critical information with their limited learning abilities; (2) for information loss during forward propagation in BNNs, IR$^2$Net leverages the information recovery method to fuse the shallow feature information with the final feature information before the classifier (or other task-specific modules) to fix the information loss and improve the accuracy.\\
\indent
With the abovementioned designs, the proposed IR$^2$Net can effectively force BNNs to focus on important information, defend against information loss in forward propagation, and then achieve advanced performance and a good trade-off between accuracy and efficiency on various networks and datasets.\\
\indent
The main contributions can be summarized as follows.\\
\indent
1) We propose IR$^2$Net, the first to mitigate the information loss and the mismatch between learning ability and information quantity from the perspective of the limited representational capability of BNNs caused by quantization.\\
\indent
2) An information restriction method is designed to restrict the input information by the generated attention masks so that the amount of input information matches the learning ability of the network, and then the representational capability of the network is fully utilized without introducing additional costs.\\
\indent
3) An information recovery method is proposed to resist the information loss in forward propagation by fusing shallow and deep information; a compact information recovery method is also proposed to reduce additional computational cost and empower the network to trade-off accuracy and computational complexity.\\
\indent
4) Extensive experimental evaluations demonstrate that the proposed IR$^2$Net achieves new state-of-the-art performance on both CIFRA-10 and ImageNet, and also has good versatility.
\begin{figure*}[!t]
	\centering
	\includegraphics[width=7in]{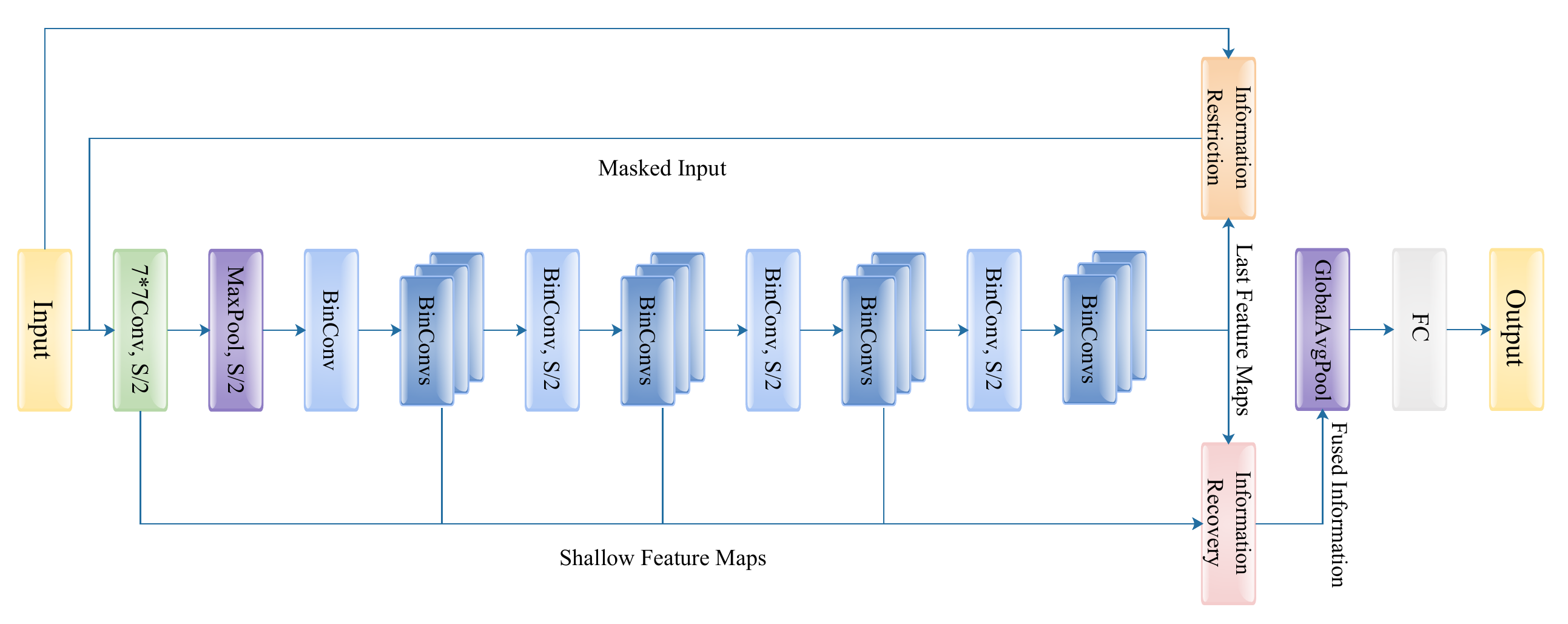}
	\caption{Illustration of IR$^2$Net. ResNet-18 is used as an example backbone. Batch-normalization layer (BN), nonlinear layer, and shortcut are omitted for simplicity. The output feature maps before downsampling layers and of the penultimate layer are selected for information recovery to retain more information with less computational cost.}
\end{figure*}

\section{Related Work}
\subsection{Network Binarization}
The pioneering study of network binarization dates back to BNN \cite{29:nips/HubaraCSEB16}, which obtains comparable accuracy on small datasets (including MNIST, SVHN \cite{30:netzer2011reading}, and CIFAR-10 \cite{31:pami/TorralbaFF08}), yet encounters severe performance degradation while on large-scale datasets (e.g., ImageNet \cite{32:cvpr/DengDSLL009}). Therefore, substantial research efforts are invested in minimizing the accuracy gap between BNNs and full-precision ones. The Enhancement of BNNs usually requires the introduction of additional computational effort. Some works focus on using a fractional amount of real-valued operations in exchange for significant accuracy gains. For instance, XNOR-Net \cite{22:eccv/RastegariORF16} improves the performance of BNNs on ImageNet to some extent by introducing real-valued scaling factors. XNOR-Net++ \cite{33:bmvc/BulatT19} on top of this by fusing the separated weights and activation scaling factors into one, which is learned discriminatively via backpropagation. Bi-Real Net \cite{23:eccv/LiuWLYLC18} connects the real-valued activation of adjacent layers to enhance the network representational capability. BBG \cite{28:icassp/ShenLGH20} adds a gated module to the connection. Real-to-Bin \cite{34:iclr/MartinezYBT20} obtains the activation scaling factors via SE \cite{35:cvpr/HuSS18}. RBNN \cite{36:nips/LinJX00WHL20} further reduces the quantization error from the perspective of intrinsic angular bias. Whereas some other works relax the constraints on the additional computational complexity for higher accuracy. ABC-Net \cite{37:nips/LinZP17} uses linear combinations of multiple binary bases to approximate the real-valued weights and activations. HORQ-Net \cite{38:iccv/LiNZY017} reduces the residual between real-valued activations and binary activations by utilizing a high-order approximation scheme. CBCN \cite{39:cvpr/LiuDXZGLJD19} enhances the diversity of intermediate feature maps by rotating the weight matrix. MeliusNet \cite{25:corr/abs-2001-05936} designs Dense Block and Improvement Block to improve the feature capability and quality, respectively. Group-Net \cite{24:cvpr/ZhuangSTL019} and BENN \cite{40:cvpr/ZhuDS19} use multiple BNNs for combination or ensemble to obtain significant improvement.\\
\indent
Although great progress has been made in the research of BNNs, the existing methods either remain a significant accuracy gap compared with full-precision networks, or introduce a large amount of computation for comparable performance, which largely offsets the advantages in compression and acceleration and deviates from the original purpose of network binarization. Therefore, IR$^2$Net is proposed, aiming at acquiring higher network accuracy with less computational complexity. Moreover, the trade-off between accuracy and efficiency is pursued by adjusting the hyperparameters introduced in IR$^2$Net, i.e., to achieve better accuracy with comparable computational cost, or to obtain comparable accuracy with less computation complexity.

\subsection{Efficient Architecture Design}
The main point of this line is to design compact architecture for model compression and acceleration. AlexNet \cite{1:nips/KrizhevskySH12} introduces group convolution to overcome the GPU memory constraints by partitioning input feature channels into mutually exclusive groups for convolution independently. However, group operation blocks the information interaction between different groups, so ShuffleNet \cite{12:cvpr/ZhangZLS18} introduces channel shuffle operation on top of group convolution to maintain the connections between groups. IGCNets \cite{41:iccv/ZhangQ0W17} uses two successive interleaved group convolutions to achieve complementarity. Xception \cite{42:cvpr/Chollet17} proposes a depth-separable convolution, which factorizes a standard convolution into depthwise convolution and pointwise convolution. MobileNet \cite{11:corr/HowardZCKWWAA17} uses depth-separable convolution to lighten the network. Based on the similarity between feature maps, GhostNet \cite{13:cvpr/HanW0GXX20} introduces the Ghost module to replace the conventional convolution to build compact neural networks. The approach along this line is orthogonal to the binarization method, whereas inspired by the lightweight structure design, we propose the compact information recovery method to empower BNNs with the ability to trade-off accuracy and efficiency while reducing the extra computational cost.

\section{Preliminaries}
In full-precision convolutional neural networks, the basic operation can be formalized as:
\begin{equation}
	z = {\omega _r} \otimes {A_r}
\end{equation}
where ${\omega _r}$ indicates the real-valued weight, ${A_r}$ is the real-valued input activation, and $\otimes$ the real-valued convolution. During the inference, the real-valued convolution operation contains a large number of floating-point operations and is computationally intensive. Network binarization aims to represent weights and activations with only 1-bit. By constraining the weights and activations to \{-1, +1\}, the convolution operations can be implemented using efficient XNOR and bitcount, which is given as follows:
\begin{equation}
	\begin{array}{c}
		{\omega _b} = sign({\omega _r}),\;{A_b} = sign({A_r})\\
		z = {\omega _b} \oplus {A_b}
	\end{array}
\end{equation}
where ${\omega _b}$ and ${A_b}$ denote the binary weight and input activation, respectively, and $\oplus$ the binary convolution. $sign( \cdot )$ is the binarization function, which is used to convert the real-valued weights and activations into binary ones, and the function takes the form as:
\begin{equation}
	sign(x) = \left\{ \begin{array}{l}
		+ 1,\;\;if\;x \ge 0\\
		- 1,\;\;otherwise
	\end{array} \right.
\end{equation}
\indent
Usually, binarization causes performance degradation and most methods \cite{22:eccv/RastegariORF16,23:eccv/LiuWLYLC18,33:bmvc/BulatT19,34:iclr/MartinezYBT20,36:nips/LinJX00WHL20,43:eccv/LiuSSC20} introduce real-valued scaling factors to reduce the quantization error and the binary convolution operation is replaced as:
\begin{equation}
	z = \alpha \beta ({\omega _b} \oplus {A_b})
\end{equation}
where $\alpha$ and $\beta$ are the scaling factors for the weights and activations, respectively (which may not be used simultaneously). Unlike these methods, in this paper, considering the property of the limited representational capability, we optimize BNNs via information restriction and information recovery, so that the scaling factors can be safely removed (although they could also be retained for compatibility with existing optimization methods).
\begin{figure*}[!t]
	\centering
	\includegraphics[width=6in]{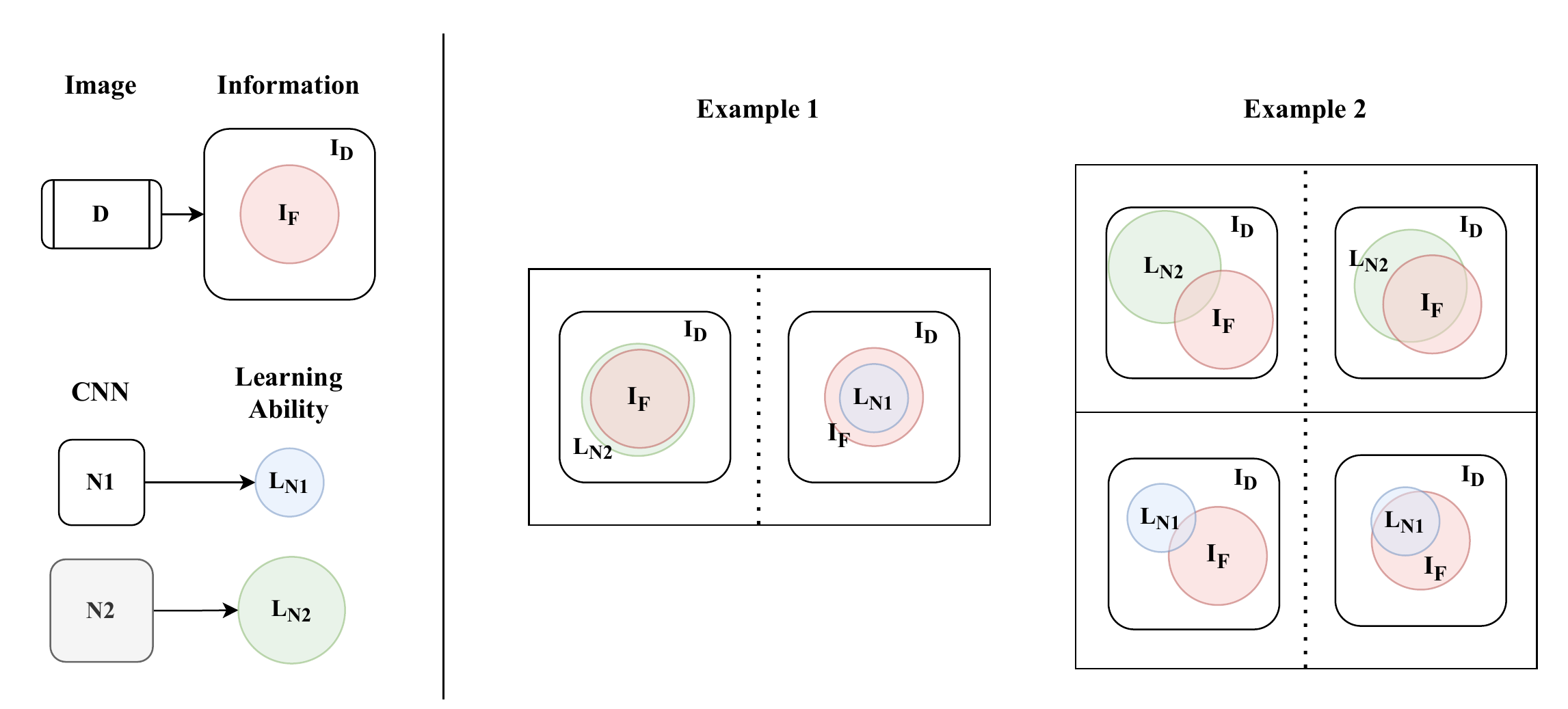}
	\caption{Rationale for the information restriction method.}
\end{figure*}

\section{Method}
In this section, we present the proposed Information Restriction and Information Recovery Network (IR$^2$Net) for network binarization. IR$^2$Net improves BNNs by tuning both the input and output of the network. Specifically, to restrict the input to induce the maximum match between the amount of input information and the learning ability of the network, and to augment the output before classifier to resist information loss in forward propagation, and then the both work together to boost the performance of BNNs.\\
\indent
An overview of the proposed IR$^2$Net is illustrated in Figure 2. IR$^2$Net is composed of two methodologies, information restriction and information recovery, for matching learning ability and resisting information loss. Specifically, the information restriction method evaluates the learning ability of the network based on the output feature maps of the penultimate layer, analyzes the learned knowledge that the network can acquire from current input, and discards some information in each sample that it cannot pay attention to, and achieves the matching between the amount of input information and the learning ability; while the information recovery method takes the penultimate layer outputs as the primary information and re-extracts the shallow feature maps as the supplementary information, then counteracts the information loss during propagation by fusing the primary information with the re-extracted supplementary information. The details of these methods are elaborated on below.

\subsection{Information Restriction}
The Information Restriction method (IRes) is motivated by the intuitive assumption that the learning ability needs to match the amount of input information that needs to be learned. As shown in Figure 3, assuming that the information contained in an image $D$ is ${I_D} = 50$, the feature information about the object that needs to be classified ${I_F} = 25$, and the network accuracy also benefit from the redundant information to some extent \cite{13:cvpr/HanW0GXX20}, thus the minimal learning ability of a network capable of accurate classification ${L_{\min }} \ge {I_F}$. On the one hand, if the learning ability of a network ${L_N} \ge {L_{\min }}$, the network is theoretically capable of accurately classifying $D$; whereas if the learning ability of a network ${L_N} < {L_{\min }}$, the amount of the feature information exceeds the learning ability of the network, it can only classify correctly with a certain probability. On the other hand, under a certain learning ability ${L_N} < {I_D}$, if the region of interest of the network (the region covered by blue and green circles in Figure 3) deviates from the region of feature information (the region covered by pink circles in Figure 3), the further the deviation, the worse the network performance, and vice versa; besides, if the larger the gap between the learning ability and the amount of the feature information, the higher the possibility of the deviation.\\
\begin{figure}[!t]
	\centering
	\includegraphics[width=1.0\linewidth]{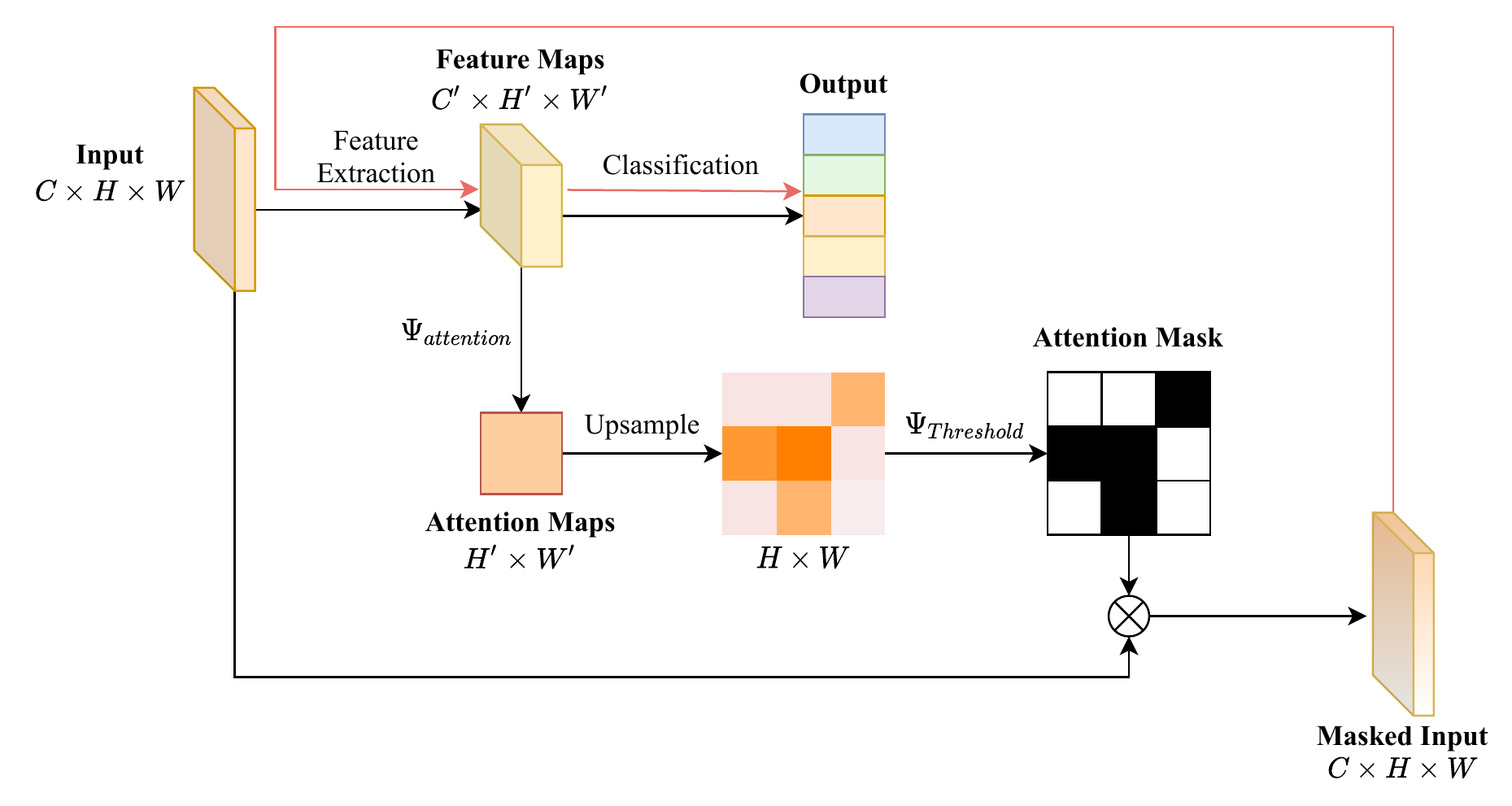}
	\caption{Details of the information restriction method, $C$, $H$, and $W$ denote the number of channels, height, and width, respectively, with super/subscripts for varying values.}
\end{figure}
\indent
The spatial location of feature information varies from image to image, so static information restriction or manual annotation is not appropriate. Usually, in CNNs, the network uses the stack of convolutional blocks as a feature extractor to extract the features of the input image, and the last linear layer as a classifier to classify the input image with the extracted features to accomplish the classification. Therefore, it is reasonable to analyze the knowledge learned by the network based on the outputs of the extractor. Specifically, as shown in Figure 4, we use the attention map ${F_A}$ generated from the basis of the output feature maps of the penultimate layer ${A_l}$ as the knowledge learned by the network:
\begin{equation}
	\begin{array}{l}
		{F_A} = {\Psi _{attention}}({A_l})\\
		{\Psi _{attention}}( \cdot ) = \sum\nolimits_{i = 1}^{C'} {|\; \cdot \;{|^2}} 
	\end{array}
\end{equation}
The generated attention map first performs bilinear upsampling to make its spatial dimension the same as the input image:
\begin{equation}
	{F_A'} = UpSample({F_A})
\end{equation}
The value of each element in the attention map represents the attention level of the network to that pixel in the input image. By setting a threshold $\tau$, we put the value of the elements with lower attention levels to 0 and the higher ones to 1 to generate an attention mask ${F_m}$ that masks the input image $D$ to achieve information restriction, as follows:
\begin{equation}
	\begin{array}{l}
		{F_m} = {\Psi _{Threshold}}(F_A')\\
		{D_m} = {F_m} \odot D
	\end{array}
\end{equation}
where $\odot$ denotes Hadamard product, ${D_m}$ is the masked image. ${\Psi _{Threshold}}$ is used to generate the mask matrices, expressed as:
\begin{equation}
	{\Psi _{Threshold}}(x){\rm{ = }}\left\{ \begin{array}{l}
		1,\;\;\;x \ge \tau \\
		0,\;\;otherwise
	\end{array} \right.
\end{equation}
\indent
It is worth noting that since the input data is variable, the range of values of the generated attention maps also varies. Therefore, the product of the mean value of the attention map and the hyperparameter $\lambda  \in [0,1]$ is used as a threshold to avoid it being out of a reasonable range:
\begin{equation}
	\tau  = \lambda  \times Mean(F_A')
\end{equation}
In addition, since the generation of the attention mask requires prior knowledge, and obtaining the knowledge introduces extra computational complexity, thus the information restriction method is only performed in the training phase. The original image is fed into the network first to obtain $Los{s_{orginal}}$ and an attention mask, and then the attention mask and the original image are used to generate the masked image which is fed into the network again to evaluate $Los{s_{masked}}$. $Los{s_{masked}}$ is used as a regularization term to merge with $Los{s_{orginal}}$ to obtain $Los{s_{total}}$ for backpropagation, thus to force the network to focus on the critical information within its limited learning ability (i.e., to improve the overlap of the regions between the interest of the network and the object feature information as Example 1 in Figure 3) without any negative impact on the model inference delay. The final loss function is defined as:
\begin{equation}
	Los{s_{total}} = \mu Los{s_{orginal}} + (1 - \mu )Los{s_{masked}}
\end{equation}
where $\mu  \in [0,1]$ is a trade-off coefficient to balance the two losses, which is set to $\mu  = 0.5$ in all experiments of this paper. The specific workflow of the information restriction method is summarized in Algorithm 1.

\begin{algorithm}[!ht]
	\caption{The workflow of IRes.}
	\textbf{Input}: image dataset $S$, initial network $N$.\\
	\textbf{Output}:the trained binary network ${N_B}$.\\
	\textbf{Training}: \\
	\quad Split the dataset $S$ into mini-batch $\{ {b_1},{b_2},...,{b_n}\}$.\\
	\quad Calculate the original loss $Los{s_{orginal}}$: \\
	\quad\quad $\begin{array}{l}
		output,{A_l} = N({b_i})\\
		Los{s_{orginal}} = criterion(output,\;{\rm{target)}}
	\end{array}$ \\
	\quad Calculate $Los{s_{masked}}$:\\
	\quad\quad According to equations (5), (6), and (7), each image in ${b_i}$ is masked to generate $b_i^m$;\\
	\quad\quad $\begin{array}{l}
		output = N(b_i^m)\\
		Los{s_{masked}} = criterion(output,\;{\rm{target)}}
	\end{array}$ \\
	\quad Calculate $Los{s_{total}}$ according to equation (10).\\
	\quad Perform backpropagation based on $Los{s_{total}}$, and update the network ${N_B} = Uptate(N)$.\\
	\quad Repeat the above process until the training is finished and return ${N_B}$.\\
	\textbf{Inference} \\
	\quad Calculate the output based on the input image $D$:\\
	\quad\quad $output = {N_B}(D)$
\end{algorithm}
\begin{figure}[!t]
	\centering
	\includegraphics[width=3.5in]{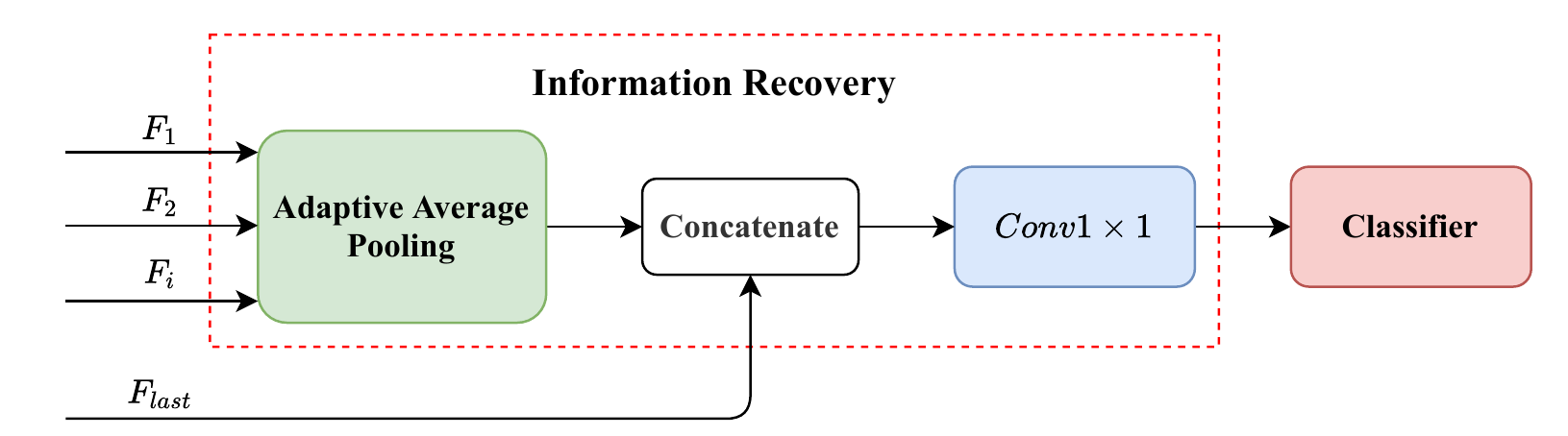}
	\caption{Details of the information recovery method, BN and nonlinear layers after the convolution are omitted.}
\end{figure}
\begin{figure*}[!t]
	\centering
	\includegraphics[width=6in]{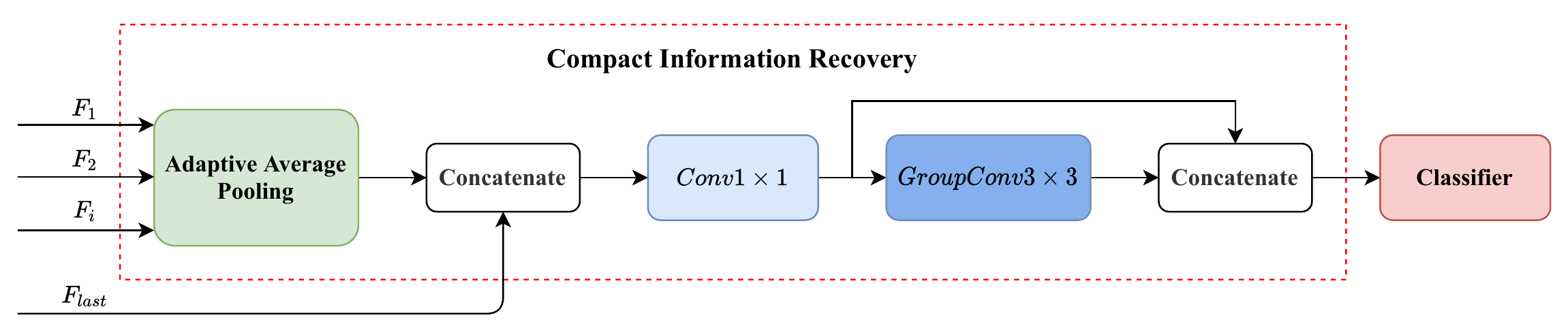}
	\caption{Details of the compact information recovery method, BN and nonlinear layers after the convolution are omitted.}
\end{figure*}
\subsection{Information Recovery}
The information restriction method can effectively increase the overlap of the regions between the interest of the network and the feature information hence improve the network performance. However, when the learning ability of the network is exceedingly limited, the knowledge learned by the network may not be sufficient to perform classification (or other tasks) effectively even if the regions overlap highly; besides, the network suffers from severe information loss in forward propagation when binarized, which also impairs the accuracy. Therefore, enhancing the learning ability of the network and fixing its information loss is essential. The Information Recovery method (IRec) enhances the representational capability by fusing multi-level feature information to improve the output diversity; meanwhile, the shallow feature information as supplementary information is fused with the output feature maps of the penultimate layer to compensate for the information loss during the propagation. The details of the information recovery method are presented in Figure 5.\\
\indent
The output feature maps of the penultimate layer ${A_l}$ are used as the primary information ${F_{last}} = {A_l} \in {R^{{C_n} \times {H_n} \times {W_n}}}$, and the shallow feature maps as the supplementary information. Since selecting overmuch shallow information will introduce a large amount of computational cost, only the output feature maps of some of the layers (as shown in Figure 2) are picked as follows: 1) the output feature maps of the first convolutional layer. The existing binarization methods usually keep the first layer as real-valued, so the output feature maps of the first layer can retain more information; 2) the output feature maps of the convolutional layer before downsampling layers. Widely used network architectures usually contain only a small number of downsampling layers, which are selected to ensure the introduction of less computational complexity while avoiding the information loss caused by downsampling.\\
\indent
Additionally, the selected shallow feature information ${F_i} \in {R^{{C_i} \times {H_i} \times {W_i}}},\;i \in [1,2,...]$ is the output feature maps of different layers with varying spatial dimensions. The information recovery method uses adaptive pooling to make the spatial dimensions of ${F_i}$ and ${F_{last}}$ consistent, i.e.,
\begin{equation}
	\begin{array}{l}
		F_i' = AdaptiveAvgPool({F_i}),\;i \in [1,2,...]\\
		F_i' \in {R^{{C_i} \times {H_n} \times {W_n}}}
	\end{array}
\end{equation}
After concatenating the shallow information corrected for spatial dimension $F_i'$ with ${F_{last}}$, the information is fused using $1 \times 1$ convolution for information recovery. The concatenation operation is defined as:
\begin{equation}
	\begin{array}{l}
		{F_{cat}} = Concatenate(F_1',F_2',...,{F_{last}})\\
		{F_{cat}} \in {R^{\sum\limits_{i = 1}^n {{C_i}}  \times {H_n} \times {W_n}}}
	\end{array}
\end{equation}
and the fusion operation is as:
\begin{equation}
	\begin{array}{l}
		{F_{fused}} = NonLinear(BN(Conv1 \times 1({F_{cat}})))\\
		{F_{fused}} \in {R^{{C_n} \times {H_n} \times {W_n}}}
	\end{array}
\end{equation}
where ${F_{fused}}$ is the final fused information and $NonLinear( \cdot )$ the nonlinear activation function (e.g. Hardtanh, PReLU, etc.). Notably, the dimensionality of ${F_{fused}}$ is the same as that of ${A_l}$, so there is no need to adjust the subsequent modules of the network.

\subsection{Compact Information Recovery}
The information recovery method effectively resists the information loss during the propagation, but its use of $1 \times 1$ convolution induces a fair amount of computational complexity. To alleviate this problem, inspired by \cite{1:nips/KrizhevskySH12,12:cvpr/ZhangZLS18,35:cvpr/HuSS18}, we propose the Compact Information Recovery method (CIRec), which reduces the computational cost by group convolution and dimensionality reduction, and then the number of groups and the ratio of dimensionality reduction can be adjusted on demand to trade-off accuracy and efficiency. The compact information recovery method can be regarded as a generalized version of the information recovery method, and the details are illustrated in Figure 6.\\
\indent
The $1 \times 1$ convolution in the information recovery method can achieve effective fusion of the feature information but with a considerable computational cost. Group convolution \cite{1:nips/KrizhevskySH12} may significantly reduce the computational complexity, but the group operation hinders the information interaction between groups, which defeats the original purpose of information fusion. Channel shuffle \cite{12:cvpr/ZhangZLS18} enables effective recovery of information interactions between groups though, empirical study shows that the convolution operation can achieve better fusion. Therefore, the compact information recovery method replaces the channel shuffle with $1 \times 1$ convolution and uses two convolutions to form a bottleneck \cite{35:cvpr/HuSS18}. The first $1 \times 1$ convolution is used for channel information interaction and dimensionality reduction:
\begin{equation}
	\begin{array}{l}
		{F_{channel}} = NonLinear(BN(Conv1 \times 1({F_{cat}})))\\
		{F_{channel}} \in {R^{\frac{{{C_n}}}{r} \times {H_n} \times {W_n}}}
	\end{array}
\end{equation}
where $r$ is the reduction ratio, and the second $3 \times 3$ group convolution for spatial information interaction and dimensionality reconstruction:
\begin{equation}
	\begin{array}{l}
		{F_{spatial}} = NonLinear(BN(GroupConv3 \times 3({F_{channel}},g)))\\
		{F_{spatial}} \in {R^{({C_n} - \frac{{{C_n}}}{r}) \times {H_n} \times {W_n}}}
	\end{array}
\end{equation}
where $g$ denotes the number of groups. $r$ and $g$ are employed to jointly adjust the computational complexity, with $r$ for coarse tuning and $g$ for fine-tuning. Notably, to further save the computational cost, the compact information recovery method does not take the output of the second convolution as the final output, but obtains the fused information by concatenating the outputs of the two convolutions \cite{13:cvpr/HanW0GXX20}:
\begin{equation}
	\begin{array}{l}
		{F_{fused}} = Concatenate({F_{channel}},{F_{spatial}})\\
		{F_{fused}} \in {R^{{C_n} \times {H_n} \times {W_n}}}
	\end{array}
\end{equation}

\section{Experiments}
To evaluate the proposed methods, we carry out comprehensive experiments on the benchmarks CIFAR-10 \cite{31:pami/TorralbaFF08} and ImageNet \cite{32:cvpr/DengDSLL009}, using VGG-Small \cite{44:eccv/ZhangYYH18}, ResNet-20, and ResNet-18 \cite{7:cvpr/HeZRS16} as network backbones, respectively. Experimental results demonstrate the superiority of IR$^2$Net. In the following, the basic setup of the experiments is stated first, including an introduction to the datasets and a description of the implementation details; and then, a series of ablation experiments are conducted on CIFAR-10; finally, a comparison of our solution with some state-of-the-arts is presented in terms of performance and complexity.

\subsection{Experimental Setting}
1) Datasets \\
\indent
\textbf{CIFAR-10}: The CIFAR-10 dataset consists of 60,000 32x32 images divided into 10 categories, 50,000 of which are the training set and the remaining 10,000 are the test set.\\
\indent
\textbf{ImageNet}: Compared to CIFAR-10, ImageNet is more challenging because of its larger size and more diverse categories. There are several versions of this dataset, of which the widely used version ILSVRC12 is adopted in this paper. ILSVRC12 is divided into 1000 categories and contains about 1.2 million training images and 50,000 test images.

2) Implementation Details \\
\indent
The proposed methods perform in an end-to-end manner so that all existing training schemes for BNNs are applicable theoretically. Among the experiments, IR$^2$Net is implemented based on Pytorch with the following setup.\\
\indent
\textbf{Network structure}: VGG-Small, ResNet-20, and ResNet-18 are employed as backbones on CIFAR-10 and ResNet-18 on ImageNet, respectively. consistent with other binarization methods, all convolutional and fully-connected layers are binarized except for the first and last one of the network; for the activation function, Hardtanh is chosen when on the CIFAR-10 dataset \cite{27:cvpr/QinGLSWYS20}, and PReLU is used while on ImageNet \cite{34:iclr/MartinezYBT20,43:eccv/LiuSSC20}.\\
\indent
\textbf{Training strategy}: Since the sign function is not differentiable,  Straight-Through Estimator (STE) \cite{50:corr/BengioLC13} or its variants \cite{21:iccv/GongLJLHLYY19,23:eccv/LiuWLYLC18} are required, and the gradient approximation of Bi-Real Net \cite{23:eccv/LiuWLYLC18} is employed in this paper. For the training method, our IR$^2$Net is trained from scratch on CIFAR-10 without leveraging any pre-trained model; whereas on ImageNet, following \cite{34:iclr/MartinezYBT20,43:eccv/LiuSSC20}, the two-stage training method of \cite{45:corr/abs-1904-05868} is adopted. We mostly follow their original papers for the rest settings, if without otherwise specified. \\
\indent
\textbf{Complexity measurement}: We measure the computational complexity of the methods with the number of operations, which is calculated in line with Real-to-Bin \cite{34:iclr/MartinezYBT20}. In addition, following ReActNet \cite{43:eccv/LiuSSC20}, we count the binary operations (BOPs) and floating-point operations (FLOPs) separately, and the total operations are evaluated using $OPs = BOPs/64 + FLOPs$.

\subsection{Ablation Study}
To investigate the effectiveness of the components in the proposed IR$^2$Net, we perform ablation studies on CIFAR-10. In all these experiments, ResNet-20 with Bi-Real Net \cite{23:eccv/LiuWLYLC18} structure is used as the backbone and trained from scratch.\\
\begin{table}[!t]
	\renewcommand{\arraystretch}{1.3}
	\caption{Ablation study on CIFAR-10 for IR$^2$Net}
	\centering
	\begin{tabular}{|c|c|c|}
		\hline
		{Method} & {Bit-width(W/A)} & {Accuracy(\%)}\\
		\hline\hline
		FP & 32/32 & 90.8  \\
		Baseline & 1/1 & 85.2  \\
		IRes & 1/1 & 86.2  \\
		IRec & 1/1 & 87.5  \\
		CIRec & 1/1 & 86.9  \\
		IR$^2$Net (IRes + CIRec) & 1/1 & 87.2 \\
		\hline
	\end{tabular}
\end{table}
\begin{table}[!t]
	\renewcommand{\arraystretch}{1.3}
	\caption{Settings of hyperparameters $r$ and $g$ on CIFAR-10 (${C_I}$ is the number of input channels for the group convolution)}
	\centering
	\begin{tabular}{|p{1in}<{\centering}|p{1cm}<{\centering}|p{1cm}<{\centering}|}
		\hline
		{Backbone} & {$r$} & {$g$}\\
		\hline\hline
		VGG-Small & 32 & ${C_I}$  \\
		ResNet-20 & 4 & ${C_I}$  \\
		ResNet-18 & 20 & ${C_I}$  \\
		\hline
	\end{tabular}
\end{table}
\begin{table}
	\renewcommand{\arraystretch}{1.3}
	\caption{Accuracy comparison between different methods on CIFRA-10 (* indicates the use of Bi-Real Net structure)}
	\begin{center}
		\begin{tabular}{|c|c|c|c|}
			\hline
			Backbone & Method & Bit-width(W/A) & Accuracy(\%) \\
			\hline\hline
			\multirow{7}*{VGG-Small} & FP & 32/32 & 91.7 \\
			&XNOR-Net \cite{22:eccv/RastegariORF16} & 1/1 & 89.8 \\
			&BNN \cite{29:nips/HubaraCSEB16} & 1/1 & 89.9 \\
			&BNN-DL \cite{46:cvpr/DingCLM19} & 1/1 & 90.0 \\
			&IR-Net \cite{27:cvpr/QinGLSWYS20} & 1/1 & 90.4 \\
			&BinaryDuo \cite{47:iclr/KimK0K20} & 1/1 & 90.4 \\
			&\textbf{IR$^2$Net} & 1/1 & \textbf{91.5} \\ 
			\hline
			\multirow{6}*{ResNet-20} & FP & 32/32 & 90.8 \\
			&DSQ \cite{21:iccv/GongLJLHLYY19} & 1/1 & 84.1 \\
			&IR-Net \cite{27:cvpr/QinGLSWYS20} & 1/1 & 85.4 \\
			&\textbf{IR$^2$Net} & 1/1 & \textbf{86.3} \\
			&IR-Net* \cite{27:cvpr/QinGLSWYS20} & 1/1 & 86.5 \\
			&\textbf{IR$^2$Net}* & 1/1 & \textbf{87.2} \\ 
			\hline
			\multirow{5}*{ResNet-18} & FP & 32/32 & 93.0 \\
			&BNN-DL \cite{46:cvpr/DingCLM19} & 1/1 & 90.5 \\
			&IR-Net \cite{27:cvpr/QinGLSWYS20} & 1/1 & 91.5 \\
			&RBNN \cite{36:nips/LinJX00WHL20} & 1/1 & 92.2 \\
			&\textbf{IR$^2$Net} & 1/1 & \textbf{92.5} \\ 
			\hline
		\end{tabular}
	\end{center}
\end{table}
\begin{table}[!t]
	\renewcommand{\arraystretch}{1.3}
	\caption{Accuracy comparison between different methods on ImageNet (if not specified, ResNet-18 is used as the backbone)}
	\centering
	\begin{tabular}{|c|c|c|c|}
		\hline
		{Method} & {Bit-width(W/A)} & {Top-1(\%)} & {Top-5(\%)}\\
		\hline\hline
		FP & 32/32 & 69.3 & 89.2  \\
		\hline
		BNN \cite{29:nips/HubaraCSEB16} & 1/1 & 42.2 & -  \\
		XNOR-Net \cite{22:eccv/RastegariORF16} & 1/1 & 51.2 & 73.2 \\
		Bi-Real Net \cite{23:eccv/LiuWLYLC18} & 1/1 & 56.4 & 79.5 \\
		XNOR-Net++ \cite{33:bmvc/BulatT19} & 1/1 & 57.1 & 79.9 \\
		IR-Net \cite{27:cvpr/QinGLSWYS20} & 1/1 & 58.1 & 80.0 \\
		BGG \cite{28:icassp/ShenLGH20} & 1/1 & 59.4 & - \\
		CI-BCNN \cite{48:cvpr/WangLT0019} & 1/1 & 59.9 & 84.2 \\
		BinaryDuo \cite{47:iclr/KimK0K20} & 1/1 & 60.9 & 82.6 \\
		Real-to-Bin \cite{34:iclr/MartinezYBT20} & 1/1 & 65.4 & 86.2 \\
		ReActNet \cite{43:eccv/LiuSSC20} & 1/1 & 65.5 & - \\
		MeliusNet29/2 \cite{25:corr/abs-2001-05936} & 1/1 & 65.7 & - \\
		MeliusNet29 \cite{25:corr/abs-2001-05936} & 1/1 & 65.8 & - \\
		\hline
		BENN \cite{40:cvpr/ZhuDS19} & $(1/1) \times 6$ & 61.1 & - \\
		CBCN \cite{39:cvpr/LiuDXZGLJD19} & $(1/1) \times 4$ & 61.4 & 82.8 \\
		ABC-Net \cite{37:nips/LinZP17} & $(1/1) \times 5$ & 65.0 & 85.9 \\
		Group-Net \cite{24:cvpr/ZhuangSTL019} & $(1/1) \times 4$ & 66.3 & 86.6 \\
		\hline
		\textbf{IR$^2$Net-D} & 1/1 & 63.8 & 85.5 \\
		\textbf{IR$^2$Net-C} & 1/1 & 66.6 & 87.0 \\
		\textbf{IR$^2$Net-B} & 1/1 & 67.0 & 87.1 \\
		\textbf{IR$^2$Net-A} & 1/1 & 68.2 & 88.0 \\
		\hline
	\end{tabular}
\end{table}
\begin{table}[!t]
	\renewcommand{\arraystretch}{1.3}
	\caption{Settings of hyperparameters $r$ and $g$ on ImageNet (${C_I}$ denotes the number of input channels for the group convolution; when $r{\rm{ = }}1$, the compact information recovery method backs off to the original one and $g$ is not applicable)}
	\centering
	\begin{tabular}{|p{1in}<{\centering}|p{1cm}<{\centering}|p{1cm}<{\centering}|}
		\hline
		{Method} & {$r$} & {$g$}\\
		\hline\hline
		IR$^2$Net-A & 1 & -  \\
		IR$^2$Net-B & 2 & ${C_I}$  \\
		IR$^2$Net-C & 4 & 8  \\
		IR$^2$Net-D & 20 & ${C_I}$  \\
		\hline
	\end{tabular}
\end{table}
\begin{table*}[!t]
	\renewcommand{\arraystretch}{1.3}
	\caption{Computational complexity analysis of different methods on ImageNet (if not specified, ResNet-18 is used as the backbone)}
	\centering
	\begin{tabular}{|c|c|c|c|c|c|}
		\hline
		{Method} & {BOPs($ \times {10^9}$)} & {FLOPs($ \times {10^8}$)} & {OPs($ \times {10^8}$)} & {OPs gap($ \times {10^8}$)} & {Accuracy gap(\%)}\\
		\hline\hline
		\textbf{IR$^2$Net-D} & 1.68 & 1.48 & 1.74 & 0 & 0  \\
		BNN \cite{29:nips/HubaraCSEB16} & 1.70 & 1.31 & 1.58 & -0.16 & -21.6  \\
		XNOR-Net \cite{22:eccv/RastegariORF16} & 1.70 & 1.33 & 1.60 & -0.14 & -12.6  \\
		Bi-Real Net \cite{23:eccv/LiuWLYLC18} & 1.68 & 1.49 & 1.75 & +0.01 & -7.4  \\
		\hline
		\textbf{IR$^2$Net-C} & 1.68 & 1.55 & 1.81 & 0 & 0  \\
		Real-to-Bin \cite{34:iclr/MartinezYBT20} & 1.68 & 1.56 & 1.82 & +0.01 & -1.2 \\
		ReActNet \cite{43:eccv/LiuSSC20} & 1.68 & 1.55 & 1.81 & 0 & -1.1 \\
		\hline
		\textbf{IR$^2$Net-B} & 1.68 & 1.59 & 1.85 & 0 & 0  \\
		MeliusNet29/2 \cite{25:corr/abs-2001-05936} & - & - & 1.96 & +0.11 & -1.3  \\
		MeliusNet29 \cite{25:corr/abs-2001-05936} & - & - & 2.14 & +0.29 & -1.2  \\
		\hline
		\textbf{IR$^2$Net-A} & 1.68 & 1.70 & 1.96 & 0 & 0  \\
		FP & 0 & 18.3 & 18.3 & +16.34 & +1.1  \\
		\hline
	\end{tabular}
\end{table*}
1) Effect of information restriction \& information recovery \\
\indent
Table \uppercase\expandafter{\romannumeral1} shows the performance of each component (W/A represents the number of bits used in weight or activation quantization). As seen in the table, both the information restriction and information recovery methods work well independently and significantly improve the accuracy. Specifically, a 1\% absolute accuracy gain is obtained with the information restriction method compared to the baseline, whereas even a 2.3\% increase is achieved with the information recovery method. The possible reason for the difference in the effectiveness of the two methods is that the information restriction method is mainly used to improve the matching problem between the amount of input information and the learning ability so that the regions between the interest of the network and the feature information are aligned, whereas the information recovery method straightly enhances the representational capability of the network and alleviate the information loss in forward propagation. However, although the information recovery method significantly improves the accuracy, it introduces a high computational cost, which can be mitigated by using the compact information recovery method instead, which balances the accuracy and efficiency by adjusting the hyperparameters $r$ and $g$. Table \uppercase\expandafter{\romannumeral1} uses the setting $r{\rm{ = }}4$ and $g{\rm{ = }}{C_I}$, with ${C_I}$ denoting the number of input channels for the group convolution. $r$ and $g$ are strategically chosen as described in Section \uppercase\expandafter{\romannumeral5}-C. Finally, IR$^2$Net achieves a 2\% accuracy increase relative to the baseline using the combination of the information restriction method and the compact information recovery method, indicating that the effects of the two components can be superimposed.\\
\indent
2) Impact of hyperparameter $\lambda$ \\
\begin{figure}[!t]
	\centering
	\includegraphics[width=3.5in]{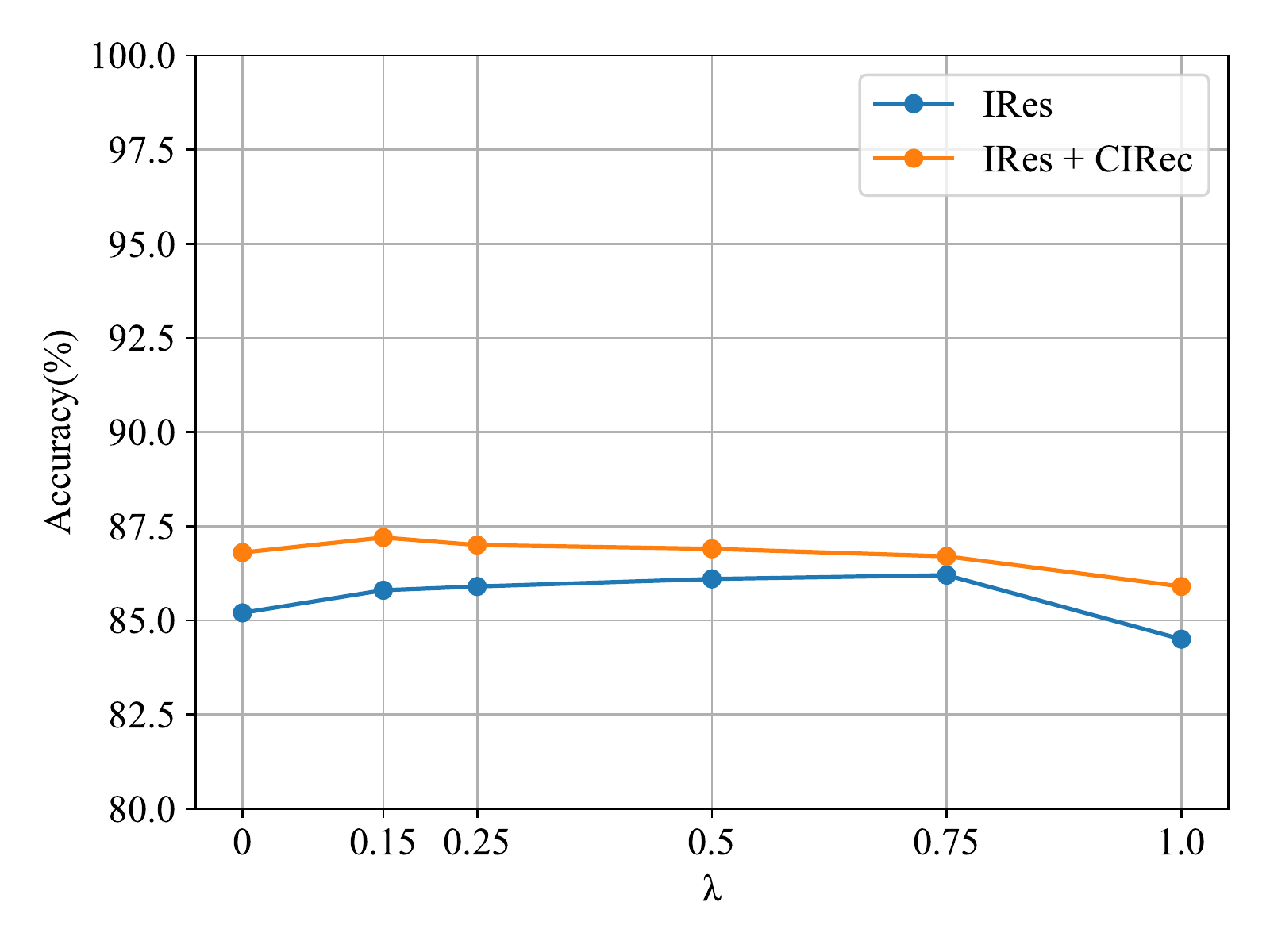}
	\caption{Impact of $\lambda$ with varying values on performance.}
\end{figure}
\begin{figure*}[!t]
	\centering
	\includegraphics[width=7in]{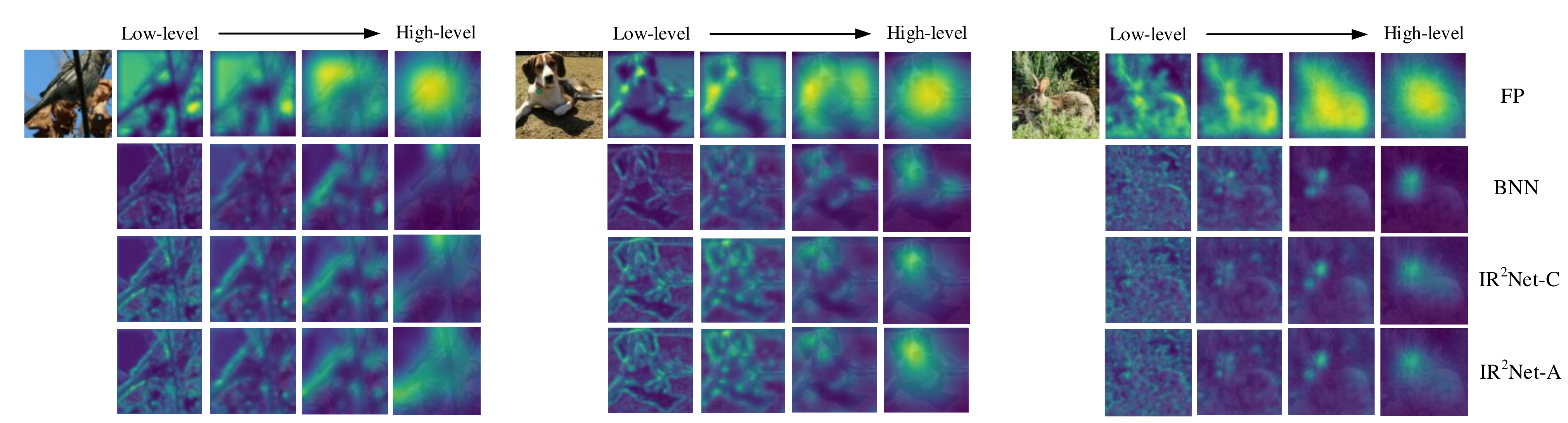}
	\caption{Comparison of the resulting attention maps.}
\end{figure*}
\indent
IR$^2$Net introduces three hyperparameters, of which $r$ and $g$ are mainly used to trade-off accuracy and efficiency on demand. In contrast, the hyperparameter $\lambda$ introduced in equation (9) is used to control the ratio of information restriction, i.e., if $\lambda {\rm{ = }}0$, it means that no information restriction is used; while the larger $\lambda$ is, the higher the restriction ratio is. Therefore, we study the impact of $\lambda$ with various values on the network accuracy, and the experimental results are plotted in Figure 7. As seen in the figure, on the one hand, when $\lambda$ is small, the accuracy is improved and with less fluctuation, compared with not using information restriction; while $\lambda$ is too large (e.g., $\lambda {\rm{ = }}1$), the accuracy decreases significantly. This indicates that the method is robust to $\lambda$ to a certain extent, but when $\lambda$ is exceedingly large, it impairs the learning of the network due to too much restriction instead. On the other hand, when $\lambda  \in [0.15,0.75]$, a larger $\lambda$ can obtain better accuracy by using only the information restriction method, whereas the opposite is true when using both the information restriction and information recovery methods. This suggests that when the information recovery method is not used, the network is with less learning ability and needs a higher information restriction ratio to match the amount of input information with the learning ability, whereas the information recovery method is used, the network has been enhanced and can accept more information, which verifies the conjecture about the relationship between learning ability and amount of input information as mentioned previously. In particular, based on the analysis of $\lambda$ with different values, we safely set $\lambda  = 0.15$ in all experiments in this paper, if not stated otherwise.

\subsection{Comparison with State-of-the-Art Methods}
We further compare the proposed IR$^2$Net with existing state-of-the-art methods on CIFAR-10 and ImageNet, respectively, to comprehensively evaluate the performance of IR$^2$Net.\\
\indent
\textbf{CIFAR-10}: On the CIFAR-10 dataset, we compare the performance of existing binarization methods with that of IR$^2$Net using VGG-Small, ResNet-20, and ResNet-18 as backbones, respectively. Noticeably, given that most existing methods use real-valued scaling factors, the FLOPs introduced by are:
\begin{equation}
	{Q_{scale}} = \sum\limits_{i = 2}^{{l_N} - 1} {{C_i} \times {H_i} \times {W_i}}
\end{equation}
where ${l_N}$ denotes the number of network layers, $i \in [2,{l_N} - 1]$, the first and last real-valued layers are excluded; ${C_i}$, ${H_i}$, and ${W_i}$ indicate the output channels, height, and width of the $i{\rm{ - th}}$ layer, respectively. And the additional FLOPs of IR$^2$Net are the sum of the computational cost of the two convolutions in Figure 6 (the FLOPs introduced by the information restriction method are zero during inference):
\begin{equation}
	\begin{array}{l}
		{Q_{CIRec}}{\rm{ = }}\frac{{{C_n}}}{r} \times {C_{in}} \times {H_n} \times {W_n} \times {K_1} \times {K_1}\\
		+ ({C_n} - \frac{{{C_n}}}{r}) \times \frac{{{C_n}}}{{gr}} \times {H_n} \times {W_n} \times {K_2} \times {K_2}
	\end{array}
\end{equation}
where ${C_{in}}$, ${C_n}$, ${H_n}$, and ${W_n}$ denote the input channels, output channels, height, and width of the compact information restriction method, respectively, and ${K_1}$, ${K_2}$ the convolution kernel size. To keep IR$^2$Net less computational cost, ensure ${Q_{CIRec}} \le {Q_{scale}}$ by adjusting $r$ and $g$, the settings are given in Table \uppercase\expandafter{\romannumeral2}. And the experimental results are listed in Table \uppercase\expandafter{\romannumeral3}, it shows that our method obtains the best accuracy on all three network backbones with large margins compared to existing methods. Particularly, over VGG-Small, the proposed method even narrows the accuracy gap between the binary model and its full-precision counterpart to 0.2\%.\\
\indent
\textbf{ImageNet}: We further investigate the performance of IR$^2$Net on ImageNet. Similar to most methods, we conduct the experiments with the ResNet-18 backbone for a fair comparison. Table \uppercase\expandafter{\romannumeral4} presents the results (the number times 1/1 indicates the multiplicative factor), where -A/B/C/D indicate different combinations of $r$ and $g$ for trading off accuracy and efficiency, the details of which are provided in Table \uppercase\expandafter{\romannumeral5}. As seen in Table \uppercase\expandafter{\romannumeral4}, even IR$^2$Net-C outperforms the other existing methods already, while IR$^2$Net-A obtains comparable accuracy to that of the full-precision counterpart, closing the gap to 1.1\%.\\
\indent
\textbf{Visualization}: In addition, to verify the effect of IR$^2$Net on the learning ability of BNNs, we visualize the attention maps learned by IR$^2$Net. As shown in Figure 8, the information regions that IR$^2$Net can focus on are significantly improved compared to the BNN (highlighted part in each figure); whereas comparing with the full-precision network, the attention is more focused on the target although the representational capability is still weaker; also, due to the different hyperparameter settings, which result in a gap in feature diversity, there are subtle differences in regions of interest between IR$^2$Net-A and IR$^2$Net-C.

\subsection{Complexity Analysis}
Table \uppercase\expandafter{\romannumeral6} shows the computational cost of different binarization methods, where the OPs gap column and Accuracy gap column indicate the gap of Ops and Top-1 accuracy between the existing methods and ours, respectively. The computational cost of IR$^2$Net-D is slightly higher than that of BNN and XNOR-Net, but there is a huge gap in accuracy. Whereas for the other methods, IR$^2$Net can achieve significant accuracy gains with less computational cost. In particular, IR$^2$Net-A obtains comparable accuracy to that of the full-precision one with $ \sim $10x computational cost reduction.

\section{Conclusion}
In this paper, we propose IR$^2$Net, which contains two components of information restriction and information recovery, from the perspective of the limited representational capability of BNNs themselves. The information restriction method motivates the amount of input information to match the learning ability of the network, improves the overlap of the regions between the interest of the network and the feature information, and then fully utilizes the representation capability; the information recovery method fuses multi-level feature information to enhance the learning ability of the network and resists the information loss in forward propagation. Besides, a compact information recovery method is further devised to reduce the computational cost and trade-off accuracy and efficiency. Experiments with various network structures on CIFAR-10 and ImageNet demonstrate the superiority of our approach.


%





\ifCLASSOPTIONcaptionsoff
  \newpage
\fi



%

\bibliographystyle{IEEEtran}
\bibliography{IEEEabrv,mybibfile}

\end{document}